# A Case for Rejection in Low Resource ML Deployment

Jerome White, Pulkit Madaan, Nikhil Shenoy, Apoorv Agnihotri, Makkunda Sharma, Jigar Doshi
agri-ai@wadhwaniai.org
Wadhwani Institute for Artificial Intelligence

## ABSTRACT

Building reliable AI decision support systems requires a robust set of data on which to train models; both with respect to quantity and diversity. Obtaining such datasets can be difficult in resource limited settings, or for applications in early stages of deployment. Sample rejection is one way to work around this challenge, however much of the existing work in this area is ill-suited for such scenarios. This paper substantiates that position and proposes a simple solution as a proof of concept baseline.

## 1 INTRODUCTION

A challenge when building models trained on small data is deciding when to deploy. Engineers must resolve whether their training set represents sufficient variation for the model to perform robustly in the deployed setting. An obvious pathway to robustness is careful collection and annotation of larger amounts of real world data. In many cases, however, that is impractical; many low resource settings, for example [5, 19]. Instead the best way forward is to deploy an imperfect model and actively monitor failure modes. This can be a delicate situation when the model is involved in human decision making. Model failures can reduce users trust and hamper the data collection process.

To address the trade-off between model improvement and user satisfaction we introduce *sample rejection* into our models inference process. Our proposal is that during deployment the model rejects samples for which it has low confidence. The sample is still collected to improve the dataset, but there is no recommendation provided to the user. While the concept of sample rejection is not new, its attention within low resource deployment settings has received relatively little attention. This paper argues that such a niche exists, and details a simple solution that serves as an example and a baseline.

## 2 MOTIVATION

This work is motivated by our deployment experience of an AI-backed mobile application (app) that assists smallholder cotton farmers with pest management decisions [4]. The app allows farmers to take photos of pests caught in pheromone traps around their field. An object detection model within the app recognizes pests in those photos to estimate the level of infestation present. Farmers are then provided with a suggestion for how to proceed based on that estimate. For many of the small holder farmers with whom we work, a seasons cotton crop represents a significant contribution to their livelihood. They take alarms about their fields quite seriously and have very little patience for suggestions that prove to be unreliable.

### 2.1 Deployment challenges

Deployment has presented many of the challenges mentioned earlier. The app was developed in early 2019 and has subsequently been used by cotton farmers across India each cotton season since.[1] Prior to the first season there were no datasets from which an object detection system could have been built. Although similar work had been explored [1, 6, 18], it had been done using private data. Efforts like Wu et al. [20] had yet to be developed. We spent the season prior to 2019 undertaking structured data collection. The outcome was a relatively small dataset with very little variance. The deployed app thus acted in part as a scaled data collection effort.

As the number of users increased, so did the variance in the images submitted. Users have always been instructed to empty traps onto a plain white sheet of paper, then to submit photos where the paper fills the frame. These instructions were easy to enforce when our user base was small. Things became more difficult as that changed. Later seasons saw photos that were clearly user experiments—photos of bikes or entire cotton fields, for example—along with others situated closer to the boundary of what we considered acceptable. Images that had pests on something other than a white sheet of paper, which were anything from a plastic tarp to a sheet of paper soiled from frequent in-field use. There were also cases where no pests were present, either because the trap did not capture them or because image quality was too low to discern. These cases have been difficult to accurately annotate, as we have found from measured experience with annotator disagreement. This in turn has made tackling the problem through supervision suboptimal.

Our deployments have also had to take into account the operating constraints of our users. Many of the farms are in remote places where internet connectivity is spotty, and many users are conscious of app storage footprints. This forces us to develop models small enough to fit within app size budgets (10's of megabytes), and limits the frequency with which we can perform model releases. These conditions make overly complicated solutions unattractive, and add importance to having confidence in model robustness at the season start.

### 2.2 Model degradation

We quantify the nature of our challenge by looking at model performance across seasons using different subsets of our data. This can be seen in Figure 1, where seasonal data splits are presented chronologically along the $x$-axis and their respective performance along the $y$-axis. We focus on mean absolute error (MAE) because accurate counting is our primary goal. K18$_p$ represents our data collection season. Seasons thereafter are prefixed with their season type and respective year.

[1] There are two predominant cotton seasons in India each year: the "summer" season lasting from approximately March through June, and the more significant "kharif" season lasting from approximately July through December.

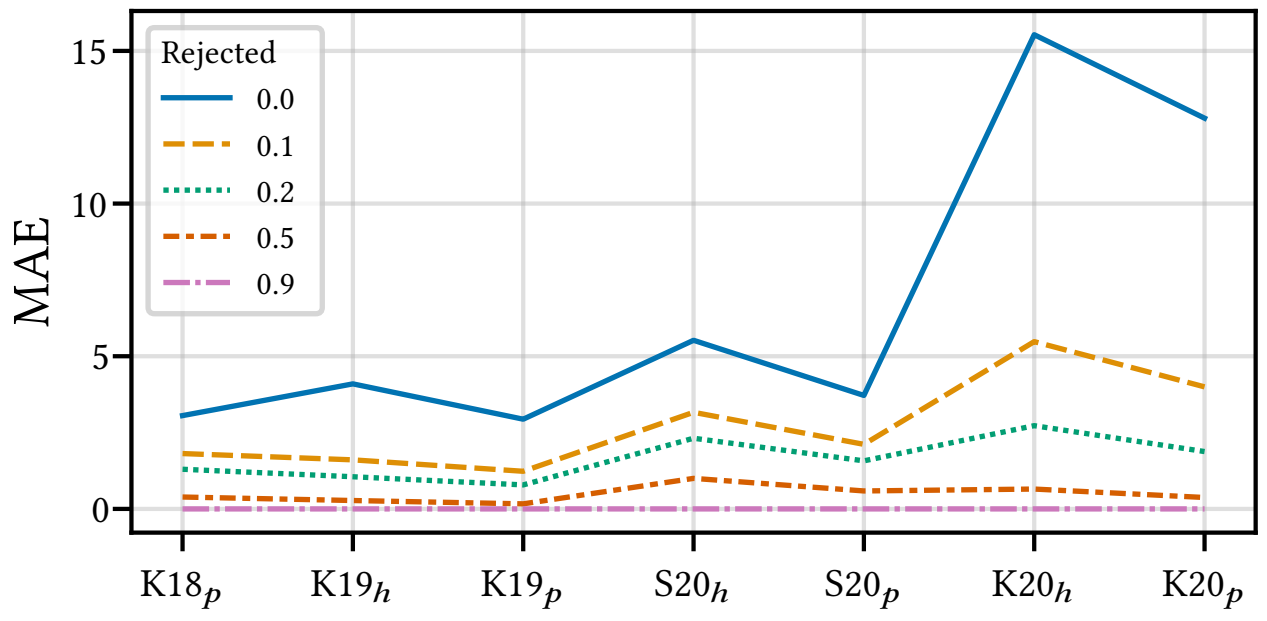

**Figure 1: Model performance across seasonal (2018–2020) development and test sets. With each season performance decreases (increasing $X_p$'s), but can be improved by training on that seasons data (decrease in immediate $X_h$'s). Performance can improve further by rejecting some fraction of worst-performing samples.**

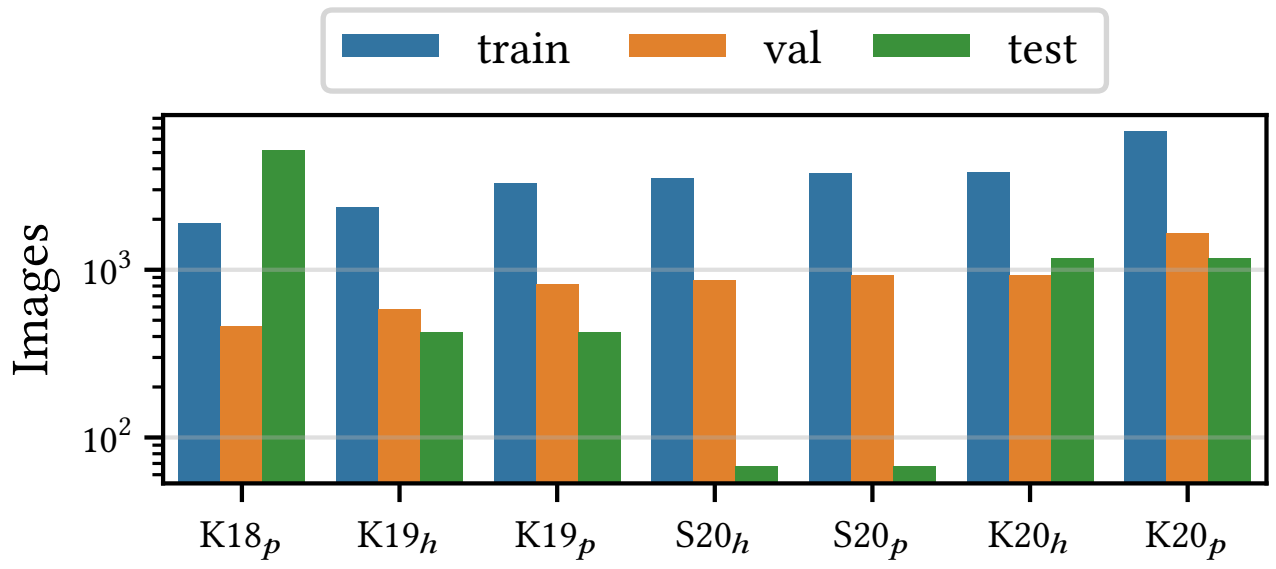

**Figure 2: Number of images in respective datasets. $h$-sets contain images from seasons prior; $p$-sets contain images from that season and seasons prior.**

The $h$ and $p$ subscript delineates the nature of that seasons development set. Each season was divided into a development and a test set, with the development set being further subdivided into training and validation. Figure 2 presents sizes for each. For sets spanning the same season and year—K19$_p$ and K19$_h$, for example—the test sets were the same, consisting of images captured during that season. The subscript differentiates where images in development sets came from: $h$-sets (*historic*) include all images captured prior-to that season; $p$-sets (*present-aware*) include all prior seasons along with images from that season.

Single Shot MultiBox Detector (SSD) [15] models were trained over each seasonal split. The solid curve in Figure 1 (rejected 0.0) tracks performance across seasons. Performance on historic sets decreases over time. This is likely due to disparities between the images seen in that season and those seen prior. When the train sets include images from that season, performance always improves. This is evident by the decrease between corresponding historic and present-aware sets.

The other curves in Figure 1, those for which "rejected" is greater-than zero, demonstrate how models would perform by rejecting worst-performing images. The rejected value denotes the percentage of images a model is allowed to discard. It is notable that at just 10 percent, K20$_h$ can be improved by almost 66 percent. On the other hand, we can achieve near perfect MAE by rejecting 90 percent of images submitted; however this would likely present undue burden for our users. Thus, while such *oracle* rejectors are impossible to deploy, they provide the bounds by which we can measure more realistic rejection systems. They also show that some level of rejection can have a positive impact on system success.

## 3 RELATED WORK

Sample rejection has a long history in the literature [2, 10]. One of the primary areas where the concept gets attention is in out-of-distribution detection. Most efforts in that area are supervised classification methods [3, 12, 14, 17]. The primary goal in our context is accurate counting, not necessarily classification. It is possible to use classification as a means to improve counting, however as mentioned in Section 2.1, obtaining clear and unambiguous annotations for what we consider out-of-distribution is not trivial. Some methods perturb samples based on a noise distribution [14]. The technique is effective at training robust understandings of domain boundaries. Whether the probability distributions used in that work would have the same benefit in our space is something to be explored.

Hüllermeier and Waegeman [11] describe uncertainty quantification, which overlaps with the principles of rejection, but differs in its application. Uncertainty quantification is concerned with the measurement of model uncertainty, while rejection is aiming at a higher accuracy with a rejection budget; see Kompa et al. [13] for an example application in medicine. Work that focuses on budget based rejection trains a rejector to maximize performance while maintaining a controlled sample rejection rate [7–9]. This perspective is too rigid for our purposes. What is most important for us is providing accurate pest management recommendations. We are, willing to tolerate variable levels of rejection to achieve that result.

## 4 METHODOLOGY

Our goal is to use rejection to improve model MAE. The approach proposed in this section does so by rejecting images based on the distribution of box confidences from the SSD model. We define *rejection level* as a two-tuple whose values guide how box confidence distributions should be defined to make image rejection decisions. This section details how these values are determined.

For a given image, SSD produces a set of bounding boxes that contain an object of interest. Each box comes with a value between zero and one expressing the models confidence in its proposal. The first value of a rejection level denotes the lower bound on box confidences—boxes below this threshold are ignored. The second value defines a lower bound on the median of the distribution of remaining boxes. Images with medians above this value are accepted and contribute to model MAE.

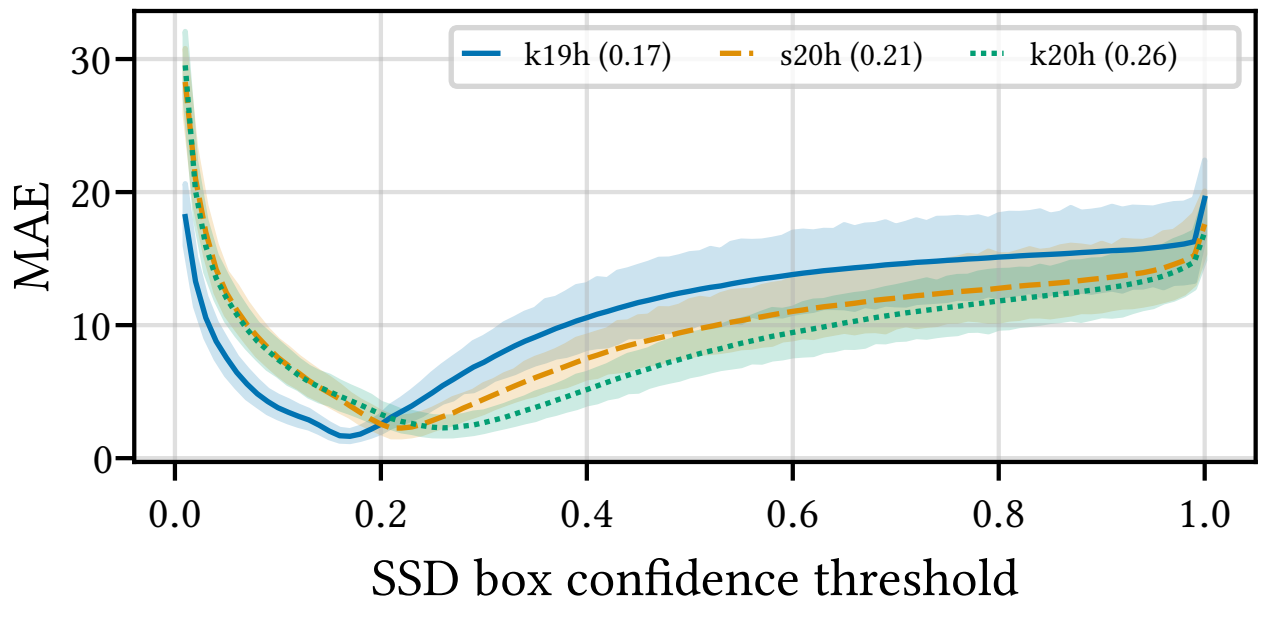

**Figure 3: Performance across seasonal validation sets as the SSD box confidence threshold is varied. Ribbons are 95 percent bootstrap confidence intervals. The legend denotes season and the respective optimal confidence threshold.**

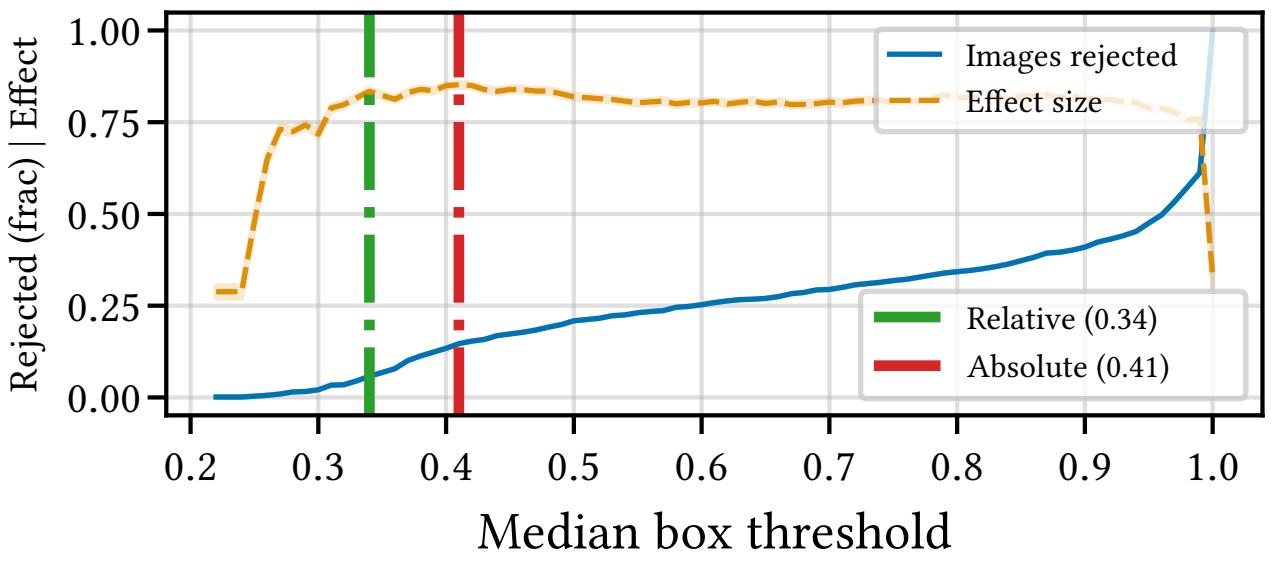

**Figure 4: Median box confidence thresholds (vertical lines) based on rejection effect size and fraction of images rejected. This plot is based on $S20_h$'s validation set, using that seasons optimal box confidence threshold (Figure 3). Dynamics for other seasons are similar.**

Our methodology uses validation sets to determine these tuples for a respective season. Tuples remain fixed during evaluation against their respective test sets (Section 5). We focus our analysis on the historic datasets as they are closest to what is experienced prior to a seasonal deployment: the ability to train on complete history, but not on images collected during the season. In this section validation sets are used for evaluation.

## 4.1 Meaningful confidence thresholds

During inference, SSD typically proposes more bounding boxes than required. To control for this disparity one of the algorithms hyperparameters is a threshold for tuning which regions are kept—boxes below the threshold are culled before final presentation. This threshold can have a substantial impact on model performance. Figure 3 presents the change in MAE ($y$-axis) for each season as the threshold is varied ($x$-axis). The optimal value for each season is presented in the figure legend. In general, thresholds that are less-than optimal are plagued by false positives; those greater-than optimal by false negatives.

Given such characteristics, we use this curve as guidance for potentially meaningful confidence distributions in a dataset. Considering all boxes—setting the confidence threshold to zero—produces a distribution that is heavily skewed toward zero. Combined with the large number of boxes produced by SSD, this acts to also skew the median calculation toward zero. We thus need a confidence threshold that produces a flatter distribution on average. We use the MAE-optimal values identified in Figure 3 as a start.

## 4.2 Choosing a median box confidence cutoff

With the box confidence fixed, we focus on determining a median by which an image rejection decisions can be made. We find this value by varying the median between zero and one. At each step we divide the images into two sets depending on whether the median of its box confidences is greater-than or equal-to the value corresponding to that step. We then calculate the absolute error (AE) for each image and look at the common language effect size [16] between the MAE's of the two partitions.

Figure 4 visualizes this process. The dashed (non-vertical) orange curve is the effect size ($y$-axis) as the median increases ($x$-axis). The effect size is calculated by sampling images from each partition, allowing us to produce a confidence interval around each threshold, represented by the ribbon around the curve. The maximum effect size corresponds to a partitioning of the images in which the probability of an image in the greater-than partition having a lower AE than an image in the less-than partition is highest. We call this point the *absolute* threshold level (Figure 4, red vertical dashed curve).

The absolute threshold does not take into account the potential number of images that may be rejected. We find a second *relative* threshold that takes that factor into consideration. The blue (solid) curve in Figure 4 tracks the number of images rejected ($y$-axis) as median increases. The relative threshold (green vertical dashed curve) is the point with the maximum effect size and minimum number of images in the rejected partition. For convenience, we use the term *global* threshold for conditions in which the absolute and relative thresholds are the same.

# 5 RESULTS

Using the methodology outlined in Section 4, we identify various rejection levels per season. This section evaluates those thresholds on respective test sets.

## 5.1 Outline

Figure 5 visualizes our results. Panels in the figure correspond to respective seasonal test sets ($x$-axis label). In a given panel each marker denotes MAE at a rejection level. A markers error bar is the 95 percent confidence interval about that mean. Rejection levels are noted in the upper legend using confidence threshold followed by median threshold. Subscripts denote whether the level is absolute ($a$), relative ($r$), or global ($g$).

Each panel also contains curves corresponding to MAE of an oracle rejector (Section 4.1). Given a fraction of images to be rejected, an oracle removes that fraction of images with the highest AE before computing the mean. There are two oracles presented. The first (solid blue curve) has access to images at a fixed confidence threshold. That threshold varies per season and is noted in the "oracle" legend. For each season, the value selected is based on the MAE-optimal identified in Figure 3. The second oracle (orange dashed curve) does not take confidence into consideration, instead selecting an images best-case AE irrespective of confidence threshold. In the case of K19$_h$, for example, that might be 0.17 for some images, and a different value for others if that value yields a better AE. The confidence-aware oracle represents a reasonable best-case baseline, as most modelers would choose a confidence threshold based on parameter tuning (Figure 3) and use that value during evaluation. As a consequence, the oracles performance at "rejected fraction" zero is thus a plausible reference. No portion of the confidence unaware oracle is achievable in reality; it is presented to give an absolute lower bound on what is possible.

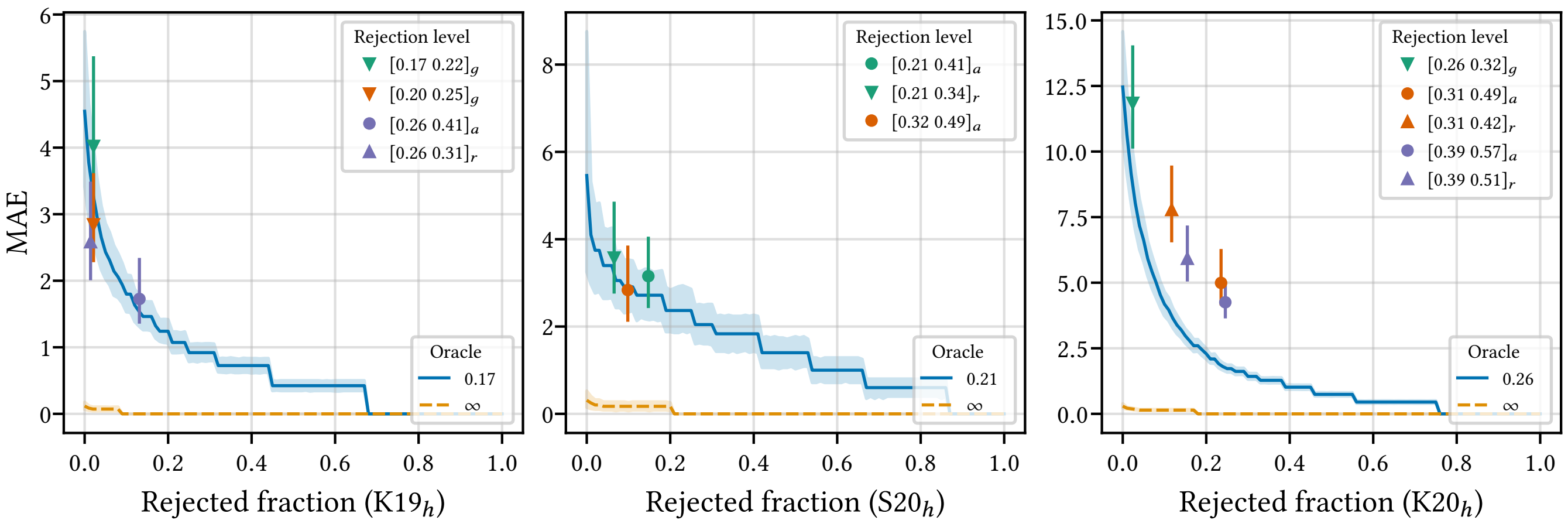

**Figure 5: Rejection levels (markers) applied to historic season test sets. Two types of oracles (curves) are presented for lower-bound reference. Error bars and ribbons denote 95 percent confidence intervals.**

## 5.2 Discussion

All rejection levels whose confidence component is optimal for its MAE ($[0.17\ 0.22]_g$ in K19$_h$, for example) perform better-than their corresponding confidence-aware oracles at the oracles zero rejected fraction. This implies that our methodology is not detrimental to model performance. Each rejection levels improvement also comes with a degree of image rejection, suggesting that our reliance on box confidence is beneficial.

In addition to MAE-optimal rejection levels, we also consider rejection levels identified using box confidence thresholds that are greater-than optimal. In Figure 3 these would be points along the $x$-axis that are greater than the respective MAE-minimums. Irrespective of season, these levels reject more images and produce better model performance than the MAE-optimal's. This is likely due to the box distribution per image being more meaningful at higher confidence thresholds.

Finally, of the two types of heuristics used for identifying median box thresholds (Section 4.2), the absolute methodology is generally better. Absolute exhibits a greater degree of rejection, however the results exhibited in Figure 5 would be acceptable for us given the MAE improvement.

# 6 CONCLUSION

Rejecting samples is a practical way of balancing model deployment with model confidence. In low resource settings, much of the existing work is impractical. The solution proposed in this paper is straightforward and has been tested on our real-world dataset. It also represents a first step in a fruitful line of work. Future exploration includes analyzing rejected images to get a better understanding of our algorithm. The methodology is global and fixed—there may be more nuance to the boxes that allow for one or both the rejection parameters to be dynamic based on the boxes of a given image. Finally, the median is effective, but rudimentary. Learning characteristics of box distributions that lead to better AEs in expectation would be ideal.